\documentclass[conference]{IEEEtran}
\IEEEoverridecommandlockouts
\usepackage[T1]{fontenc}
\usepackage{amsmath}
\usepackage{graphicx,verbatim}
\usepackage{orcidlink}
\usepackage{booktabs}
\usepackage{graphicx}
\usepackage{multirow}
\usepackage{booktabs}
\usepackage{adjustbox}
\usepackage{float}
\usepackage{comment}
\usepackage{cite}
\usepackage{amsmath,amssymb,amsfonts}
\usepackage{algorithmic}
\usepackage{graphicx}
\usepackage{textcomp}
\usepackage{xcolor}
\usepackage{fancyhdr}

\makeatletter
\newcommand{\linebreakand}{%
  \end{@IEEEauthorhalign}
  \hfill\mbox{}\par
  \mbox{}\hfill\begin{@IEEEauthorhalign}
}
\makeatother

\def\BibTeX{{\rm B\kern-.05em{\sc i\kern-.025em b}\kern-.08em
    T\kern-.1667em\lower.7ex\hbox{E}\kern-.125emX}}
\begin{document}

\title{A General Framework for Self-Supervised Denoising with Patch Aggregation\\

}

\author{%
\IEEEauthorblockN{Huy Minh Nhat Nguyen}
\IEEEauthorblockA{
\textit{Department of Computer Science and Engineering}\\
\textit{Vietnamese German University}\\
Ho Chi Minh City, Vietnam\\
10423045@student.vgu.edu.vn}
\and
\IEEEauthorblockN{Triet Hoang Minh Dao}
\IEEEauthorblockA{
\textit{Department of Computer Science and Engineering}\\
\textit{Vietnamese German University}\\
Ho Chi Minh City, Vietnam\\
10423179@student.vgu.edu.vn}
\linebreakand
\IEEEauthorblockN{Chau Vinh Hoang Truong}
\IEEEauthorblockA{
\textit{Department of Computer Science and Engineering}\\
\textit{Vietnamese German University}\\
Ho Chi Minh City, Vietnam\\
16076@student.vgu.edu.vn}
\and % <-- forces a new centered ro
\IEEEauthorblockN{Cuong Tuan Nguyen}
\IEEEauthorblockA{
\textit{Department of Computer Science and Engineering}\\
\textit{Vietnamese German University}\\
Ho Chi Minh City, Vietnam\\
cuong.nt2@vgu.edu.vn}
\vspace{-1cm}
}

\maketitle

% {\footnotesize \textsuperscript{*}Note: Sub-titles are not captured for https://ieeexplore.ieee.org  and should not be used}

%copyright notice
\thispagestyle{plain}
\fancypagestyle{plain}{
\fancyhf{} % clear all header and footer fields
\fancyfoot[L]{979-8-3315-0266-9/25/\$31.00~\copyright2025~IEEE} % change copyright notice here if required
\renewcommand{\headrulewidth}{0pt}
\renewcommand{\footrulewidth}{0pt}
}

\begin{abstract}
Optical Coherence Tomography (OCT) is a widely used non-invasive imaging technique that provides detailed three-dimensional views of the retina, which are essential for the early and accurate diagnosis of ocular diseases. Consequently, OCT image analysis and processing have emerged as key research areas in biomedical imaging. However, acquiring paired datasets of clean and real-world noisy OCT images for supervised denoising models remains a formidable challenge due to intrinsic speckle noise and practical constraints in clinical imaging environments. 

To address these issues, we propose SDPA++: A General Framework for Self-Supervised Denoising with Patch Aggregation. Our novel approach leverages only noisy OCT images by first generating pseudo-ground-truth images through self-fusion and self-supervised denoising. These refined images then serve as targets to train an ensemble of denoising models using a patch-based strategy that effectively enhances image clarity. Performance improvements are validated via metrics such as Contrast-to-Noise Ratio (CNR), Mean Square Ratio (MSR), Texture Preservation (TP), and Edge Preservation (EP) on the real-world dataset from the IEEE SPS Video and Image Processing Cup. Notably, the VIP Cup dataset contains only real-world noisy OCT images without clean references, highlighting our method’s potential for improving image quality and diagnostic outcomes in clinical practice.
\end{abstract}


\section{Introduction}
\subsubsection{Background}
Optical Coherence Tomography (OCT) is a pivotal imaging technology in ophthalmology, enabling detailed cross-sectional visualization of the retina. This technique is critical for diagnosing and managing conditions such as Age-related Macular Degeneration (AMD), Diabetic Retinopathy (DR), and Diabetic Macular Edema (DME). However, OCT images are frequently affected by speckle noise and other imaging artifacts, which complicate interpretation and may lead to inaccuracies in diagnosis.

\subsubsection{Dataset Description} 
The Signal Processing Society’s 2024 Video and Image Processing Cup (VIP Cup) \cite{IEEEVIPCup2024} challenges participants to enhance OCT image quality using a dataset composed of noisy OCT B-scan images. These scans are divided into three classes: 
\begin{itemize}
    \item Normal/healthy (42 subjects)
    \item Diabetic patients with DME (30 subjects)
    \item Non-diabetic patients with ocular diseases such as AMD, CNV, or Macular Hole (MH)
\end{itemize}
Each subject has between 30 to 300 grayscale B-scan images at 300$\times$300 resolution. Combining images from all subjects yields about 18,000 noisy OCT scans for the denoising challenge. No clean “ground truth” scans are available, highlighting the need for self-supervised or unsupervised methods.

\subsubsection{Motivation} 
Traditional denoising algorithms such as non-local means \cite{non_local_means}, wavelet-based techniques \cite{wavelet}, and median filtering \cite{median_filtering} have been applied to enhance OCT image quality. While computationally straightforward, these conventional approaches often struggle to remove heavy speckle noise while preserving fine anatomical structures. Further, supervised learning-based algorithms typically require paired noisy and clean images, which are difficult to obtain in medical settings.

To overcome this limitation, self-supervised and unsupervised strategies have garnered attention. Techniques like self-fusion \cite{SelfFusion}, Blind-Spot networks \cite{BlindSpot}, Noise2Noise \cite{Noise2Noise}, Noise2Void \cite{Noise2Void}, and Noise2Self \cite{Noise2Self} aim to learn denoising directly from noisy data. However, simply applying blind-spot or neighbor-based methods to extremely noisy data can still yield suboptimal results, particularly if the noise level is high.

\subsubsection{Contribution}
We propose a novel two-stage pipeline that creates a pseudo-clean reference from noisy images alone, then uses multiple self-supervised and supervised denoising models to refine the results further. Our main contributions are: \textbf{(i) Self-fusion pre-processing:} We exploit the 3D volume nature of OCT data by fusing neighboring B-scans to remove noise. \textbf{(ii) Self-supervised refinement:} We apply a neighbor-based self-supervised method to generate pseudo-ground-truth images without requiring any clean references. \textbf{(iii) Patch-ensemble of multiple models:} We train advanced denoising models (e.g., Restormer, NAFNet, CGNET and diffusion-based models) using pseudo-clean targets, then fuse their outputs via a patch-based ensemble to achieve superior denoising quality.

Figure~\ref{fig:drawio} illustrates an overview of our self-supervised and ensemble-based denoising framework for OCT images.

\section{Related Works}
\subsection{Traditional Denoising Methods}
Conventional denoising techniques typically rely on mathematical/statistical modeling. Non-local means \cite{non_local_means} identifies and averages similar patches across an image, effectively removing noise but sometimes blurring features if the patch similarity criteria are not carefully tuned. Wavelet thresholding \cite{wavelet} decomposes images into multiple scales/frequencies, selectively attenuating noise but risking loss of fine details when thresholding is severe. Median filtering \cite{median_filtering} replaces each pixel’s intensity with the median in a local neighborhood, effectively removing outliers but often sacrificing edge sharpness. Although these methods are simple and interpretable, their performance can degrade significantly in OCT scans with heavy speckle noise or complex structures.

\subsection{Supervised Denoising Methods}
Supervised deep learning approaches rely on paired noisy-clean datasets. U-Net \cite{UNet} introduced skip connections to preserve high-resolution spatial information in medical image segmentation, and by extension, denoising. Restormer \cite{Restormer} uses a transformer-based architecture that captures long-range dependencies, improving restoration quality. NAFNet \cite{NAFNet} refines network design to handle a wide range of image restoration tasks with fewer parameters. CascadedGaze (CGNet) \cite{CGNet}, an efficient encoder-decoder architecture with a novel Global Context Extractor that captures global information without self-attention. Though these methods are powerful, they presume the availability of clean reference images, which is often unrealistic in clinical OCT data acquisition.

\subsection{Self-Supervised and Unsupervised Denoising Methods}

In the absence of clean references, self-supervised paradigms remove the need for paired data. Blind-spot methods—Noise2Void \cite{Noise2Void} and Noise2Self \cite{Noise2Self}—mask input pixels and train the network to predict them from surrounding context, preventing identity shortcuts and forgoing ground truth. Noise2Noise \cite{Noise2Noise} instead learns from paired noisy realizations of the same scene, while Neighbor2Neighbor \cite{Nei2Nei} relaxes this further by pairing randomly subsampled patches within a single image volume.

Self-fusion \cite{SelfFusion} aggregates adjacent slices/frames to form a cleaner proxy. Blind2Unblind \cite{Blind2Unblind} and AP-BSN \cite{APBSN} refine blind-spot training and account for spatially correlated noise. Generative approaches based on Denoising Diffusion Probabilistic Models (DDPMs) \cite{DDPM} learn a reverse noising process, enabling flexible modeling of complex noise distributions for high-quality restoration.
quality restoration.

\section{Methodology}
\begin{figure}[h]
    \centering
    \includegraphics[width=\linewidth]{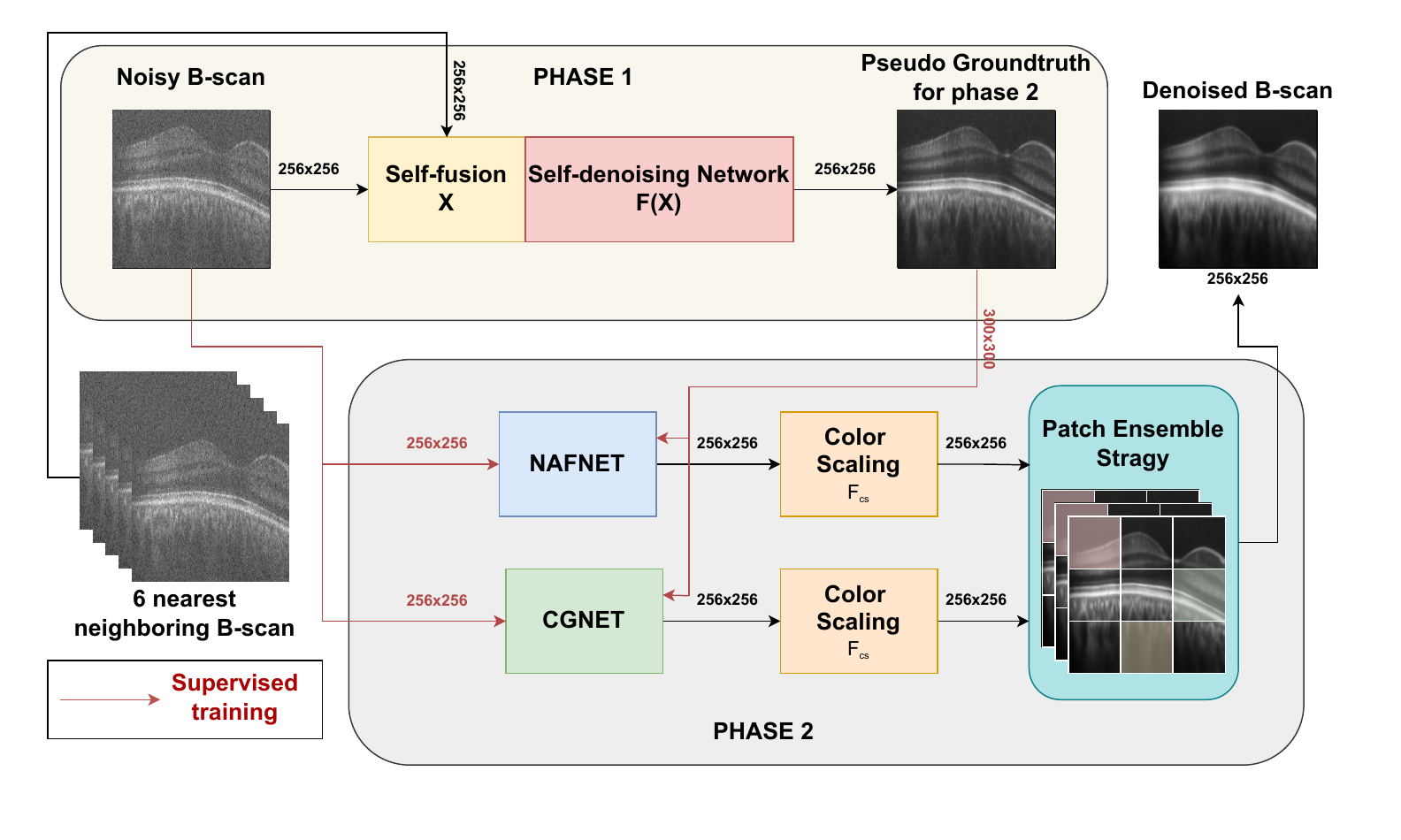}
    \vspace{-3em}
    \caption{Overview of proposed denoising method}
    \label{fig:drawio}
\end{figure}

Our denoising framework consists of two main phases (see Fig.~\ref{fig:drawio}). In \textbf{Phase 1}, we generate pseudo-clean images from the original noisy dataset using two steps: self-fusion and self-supervised denoising. In \textbf{Phase 2}, these pseudo-clean outputs become the training targets for multiple denoising models, whose results are then combined via patch-based ensemble.

\subsection{Phase 1: Self-Fusion and Self-Supervised Refinement}

\paragraph{Self-Fusion Preprocessing.} 
For each B-scan $B_i$ in an OCT volume, we collect its six nearest neighboring B-scans $B_{i-3}, \dots, B_{i+3}$ (excluding $B_i$ itself). We register them to $B_i$ if necessary (some volumes might be pre-aligned) and calculate a weighted fusion:
\[
\tilde{B}_i(x,y) = \sum_{j = i-3}^{i+3} w_j(x, y) \cdot B_j(x, y),
\]
where $w_j(x, y)$ is proportional to the patch similarity between $B_j$ and $B_i$. This produces an intermediate self-fused image that reduces random speckle noise.

\paragraph{Neighbor2Neighbor.}
Next, we apply the neighbor-based self-supervised training scheme proposed in \cite{Nei2Nei} to refine $\tilde{B}_i$. By sampling random patches within $\tilde{B}_i$ to form noisy pairs, the network learns to predict one patch from its neighbor, effectively denoising without requiring a clean ground truth. Let $f_\theta$ be our denoising network, and $(g_1(y), g_2(y))$ represent noisy patch pairs. The training loss:
\[
\mathcal{L} = \|f_\theta(g_1(y)) - g_2(y)\|_2^2 + \gamma \cdot \mathcal{L}_{\text{reg}},
\]
with a regularization $\mathcal{L}_{\text{reg}}$ that penalizes disagreement between pairs of noisy predictions, promotes consistency. After training, $f_\theta(\tilde{B}_i)$ yields a pseudo-clean image $C_i$.

\subsection{Phase 2: Supervised Denoising and Patch Ensemble}

We then treat $(B_i, C_i)$ as noisy-clean pairs to train two advanced models: NAFNet \cite{NAFNet} and CGNET \cite{CGNet}. Each model learns to map $B_i \mapsto C_i$ under supervised settings. Although $C_i$ is itself imperfect, it remains a far cleaner version than $B_i$, effectively serving as a stand-in for true ground truth.

\textbf{Patch-Based Image Ensemble: }
Once these two models (NAFNET and CGNET) are trained, each produces a denoised output for the same input $B_i$, each at a resolution of $256 \times 256$. We aim to combine the strengths of each model via a sliding-window approach:
\begin{enumerate}
    \item \textbf{Split into patches.} Split each model’s $256 \times 256$ output into overlapping patches (window of size $k \times k$, with a stride that ensures slight overlap among consecutive patches).
    \item \textbf{Local quality assessment.} At each patch location, evaluate a local measure of image quality using the combination of CNR, MSR, TP and EP metrics to gauge noise reduction and detail preservation.
    \item \textbf{Select the best patch.} Among the three patches (one from each model) covering the same spatial region, select the patch that yields the highest score on the local quality metric. 
    \item \textbf{Blend via averaging.} In overlapping regions, average the selected patches’ pixel values to create a seamless final image without boundaries or abrupt transitions.
\end{enumerate}
This procedure ensures that the locally best denoising solution among the two networks is chosen, effectively reducing noise while preserving important details.

\begin{figure}[h]
    \centering
    \includegraphics[width=1\linewidth]{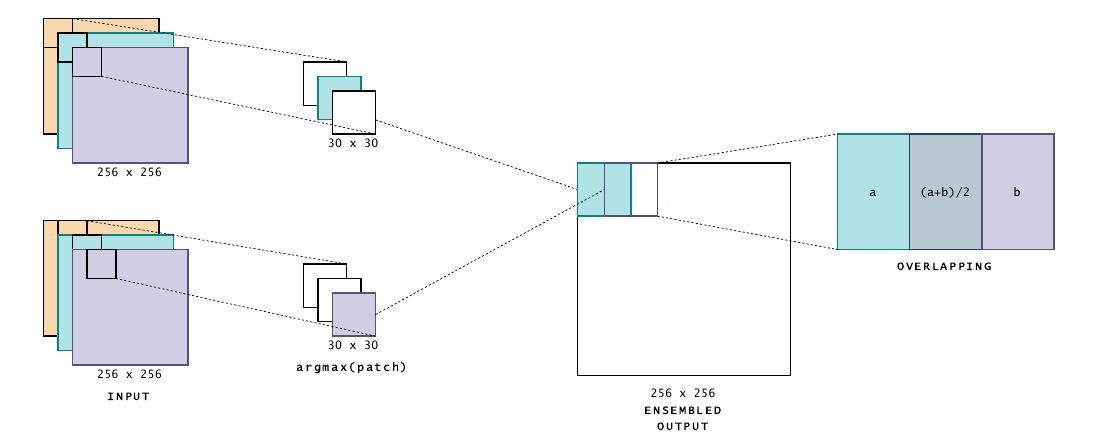}
    \vspace{-2em}
    \caption{Patch Ensemble Strategy}
    \label{fig:ensemble}
    
\end{figure}
\vspace{-2em}
\subsection{Color Scaling Post-Processing}
After ensemble fusion, we apply an intensity rescaling step to maximize contrast:
\[
I_{\text{norm}} = \frac{I}{255}, 
\quad
F_{\text{cs}}(I_{\text{norm}}) = 
\left(\frac{I_{\text{norm}} - \min(I_{\text{norm}})}{\max(I_{\text{norm}}) - \min(I_{\text{norm}})}\right)
,
\]
ensuring the darkest pixels map to 0 and the brightest map to 255 for improved visibility of retinal structures. More results in ablation study \ref{table:colortable}

\begin{figure}[h]
    \centering
    \includegraphics[width=\linewidth]{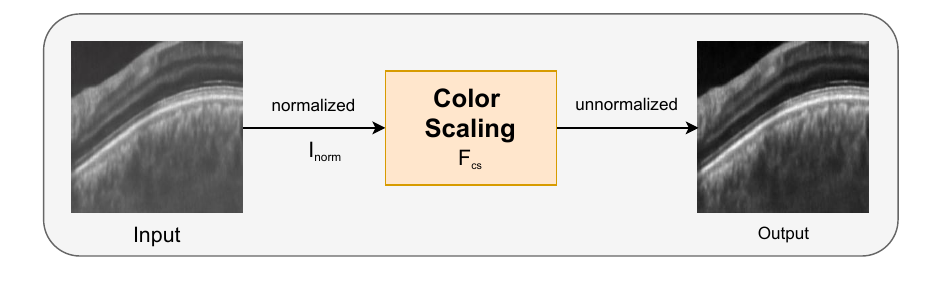}
    \vspace{-2em}
    \caption{The effect of applying the proposed contrast enhancement function on the image.}
    \label{fig:contrast}
\end{figure}

\subsection{Model Inference}
After completing Phase 1 and Phase 2 processing, the model produces a denoised OCT image. We then apply our method across the entire training dataset to obtain paired noisy and denoised images. These paired samples are used to train a new CGNET model \cite{CGNet} using a supervised learning approach. Once trained, only the CGNET model is required for inference. See more in ablation study \ref{table:ablation_methods}

\section{Experiment}
\subsection{Dataset and metrics}
In this paper, we only use the training set of the dataset provided in the Signal Processing Society’s 2024 Video and Image Processing Cup challenge. The dataset has been captured in Didavaran clinic, Isfahan, Iran, using a custom-made Swept-Source OCT imaging system built in department of Biomedical Engineering, University of Basel. The central wavelength and spectral bandwidth of the OCT system are 1064 nm and 100 nm, respectively. The training set includes 100 OCT volume OCT data from 100 subjects, each volume belongs to one of three types (healthy, diabetic with DME, and non-diabetic patients with other ocular diseases), where each volume contains several B-scans of size 300x300. We split the training set into 2 smaller sets, training, and testing set with a ratio of 9:1. Besides that, we convert all images to a size of 256x256. For patch-based ensemble, we set the window size of 16x16 pixels, with a stride of 4 pixels. 

For evaluation of the image quality as well as score calculation in ensemble strategy, we use 4 metrics Contrast-to-Noise Ratio (CNR), Mean-to-standard-deviation ratio (MSR), Texture Preservation (TP), and Edge Preservation (EP). 

\textbf{CNR}: The Contrast-to-Noise Ratio (CNR) measures the contrast between significant structures and the background relative to noise, assessing how well the denoising algorithm enhances feature visibility while reducing noise. CNR is defined as:
\[
    CNR = 10 \log \left( \frac{|\mu_{f} - \mu_{b}|}{\sqrt{0.5(\sigma_{f}^{2} + \sigma_{b}^{2})}} \right)
\]
where $\mu_{f}$ and $\sigma_{f}$ represent the mean and standard deviation of the foreground regions, and $\mu_{b}$ and $\sigma_{b}$ denote those of the background regions.

\textbf{MSR}: The Mean-to-Standard-Deviation Ratio (MSR) evaluates the clarity and sharpness of the image by comparing the signal's intensity to the noise variability, indicating the effectiveness of the noise reduction process. MSR is given by:
\[
    MSR = \frac{\mu_{f}}{\sigma_{f}}
\]
where $\mu_{f}$ and $\sigma_{f}$ are the mean and standard deviation of the foreground regions, respectively.

\textbf{TP}: Texture Preservation (TP) assesses the denoising algorithm’s ability to retain the original texture of the image, which is crucial for accurate medical interpretation. TP is evaluated using the formula:
\[
    TP = \frac{\sigma_{m}^{2}}{\left(\sigma_{m}^{'2}\right)} \sqrt{\frac{\mu_{den}}{\mu_{in}}}
\]
where $\sigma_{m}$ and $\sigma_{m}^{'}$ denote the standard deviation of the de-noised and noisy images in the m-th Region of Interest (ROI), respectively. $\mu_{den}$ and $\mu_{in}$ represent the mean values of the de-noised and noisy images, respectively. 

\textbf{EP}: Edge Preservation (EP) measures how well the denoising algorithm preserves the edges of structures within the image, ensuring that critical boundaries are maintained for accurate medical analysis. EP is calculated as:

EP =                              
\[
    \frac{\Gamma(\Delta I'_{m} - \overline{\Delta I'_{m}}, \Delta I_{m} - \overline{\Delta I_{m}})}{\sqrt{\Gamma(\Delta I'_{m} - \overline{\Delta I'_{m}}, \Delta I'_{m} - \overline{\Delta I'_{m}}) \Gamma(\Delta I_{m} - \overline{\Delta I_{m}}, \Delta I_{m} - \overline{\Delta I_{m}})}}
\]
where $I_m$ and $I'_m$ are matrices that contain the de-noised and the noisy image regions, respectively, in the m-th ROI. $\Delta$ represents the Laplacian operator and $\Gamma$ measures the correlation of images.

We validate SPDA++ on the VIP Cup dataset by splitting the 18,000 noisy B-scans into training and validation sets. Preliminary evaluations using metrics such as Contrast-to-Noise Ratio (CNR), Mean Square Ratio (MSR), Texture Preservation (TP), and Edge Preservation (EP) show that our two-stage strategy achieves notable improvements over direct blind-spot methods or single-model baselines. % Qualitative inspection by clinical experts further confirms that our outputs contain clearer retinal layers and fewer artifacts.

\subsection{Experimental Results}
First, we evaluated the performance metrics of the proposed patch-based ensemble method (Patch Ensemble) as compared with Voting Ensemble, Self-fusion, and the single models of DDPM, NAFNET, and Restormer. From the result of Table \ref{table:methods_comparison}, Patch-based ensemble outperforms voting ensemble in MSR, TP, and EP, while slightly behind in CNR. From the result, we can see the instability in the performance of the single models, which complicates the global weighting in a voting ensemble approach. Consequently, this underscores the necessity to assess individual model outputs prior to their integration in the ensemble process.

\begin{table}[H]
\vspace{-2em}
\centering
\caption{Comparison of various methods across two folds and their average. The best top 3 results are \textcolor{red}{Red}, \textcolor{blue}{Blue} and \textcolor{brown}{Brown}.}
\setlength{\tabcolsep}{4pt} % Adjust the column separation
\scalebox{0.72}{
\begin{tabular}{lcccccccccccc}
\toprule
 & \multicolumn{4}{c}{Fold 1} & \multicolumn{4}{c}{Fold 2} & \multicolumn{4}{c}{Average} \\ 
\cmidrule(lr){2-5} \cmidrule(lr){6-9} \cmidrule(lr){10-13}
 & CNR & MSR & TP & EP & CNR & MSR & TP & EP & CNR & MSR & TP & EP \\ 
\midrule
Self-fusion & 7.87 & \textcolor{blue}{3.96} & 0.91 & \textcolor{blue}{0.76} & 8.15 & \textcolor{blue}{4.29} & 0.90 & 0.71 & 8.01 & \textcolor{blue}{4.12} & 0.91 & 0.73 \\ 
Neighbor2Neighbor & 11.08 & \textcolor{red}{5.38} & 0.68 & \textcolor{red}{0.77} & 11.63 & \textcolor{red}{6.12} & 0.67 & \textcolor{red}{0.77} & 11.35 & \textcolor{red}{5.75} & 0.67 & \textcolor{red}{0.77} \\ 
DDPM & \textcolor{brown}{11.89} & 3.47 & 1.01 & 0.73 & 12.68 & 4.10 & 0.97 & \textcolor{brown}{0.74} & \textcolor{brown}{12.28} & 3.79 & 0.99 & 0.73 \\ 
NAFNET & 11.49 & 3.38 & \textcolor{red}{1.21} & 0.73 & \textcolor{brown}{12.90} & 4.25 & \textcolor{red}{1.13} & 0.73 & 12.19 & 3.81 & \textcolor{red}{1.17} & 0.73 \\ 
Restormer & 11.63 & 3.31 & 1.01 & \textcolor{brown}{0.74} & 12.18 & 3.78 & 0.95 & 0.73 & 11.90 & 3.54 & 0.98 & \textcolor{blue}{0.75} \\ 
Voting Ensemble & \textcolor{red}{11.99} & 3.24 & \textcolor{brown}{1.10} & 0.73 & \textcolor{red}{13.00} & 3.93 & \textcolor{brown}{1.04} & 0.73 & \textcolor{red}{12.49} & 3.58 & \textcolor{brown}{1.07} & 0.73 \\ 
\textbf{Patch Ensemble} & \textcolor{blue}{11.96} & \textcolor{brown}{3.53} & \textcolor{blue}{1.18} & \textcolor{brown}{0.74} & \textcolor{blue}{12.92} & \textcolor{brown}{4.28} & \textcolor{blue}{1.12} & \textcolor{blue}{0.75} & \textcolor{blue}{12.44} & \textcolor{brown}{3.90} & \textcolor{blue}{1.15} & \textcolor{brown}{0.74} \\ 
\bottomrule
\end{tabular}
}
\label{table:methods_comparison}
\end{table}

In the second experiment, we evaluated the performance of Patch Ensemble
with different configurations of patch selection criteria. Table 2 shows the results
of Patch Ensemble with the criteria based on single metrics of CNR, TP, MSR
and EP as compared with the criteria using a weighted combination of multiple
metrics (Patch Ensemble (All)). From the result, we can see that Patch Ensemble
with the criteria using a single metric enhances the performance for that specific
metric but reduces the performance of other metrics. We propose a combination
of metrics as the selection criteria. The best combination of metrics is found
experimentally: CNR*3 + MSR*2 + TP*5. With this approach, the ensemble
method achieves an effective trade-off among the evaluated metrics, thereby
optimizing overall result quality.

\begin{table}[h!]
\centering
\vspace{-1em}
\caption{Comparison of various patch ensemble scores across two folds and their average. The best top 3 results are \textcolor{red}{Red}, \textcolor{blue}{Blue}, \textcolor{brown}{Brown}}
\setlength{\tabcolsep}{4pt} % Adjust the column separation
\scalebox{0.69}{
\begin{tabular}{lcccccccccccc}
\toprule
 & \multicolumn{4}{c}{Fold 1} & \multicolumn{4}{c}{Fold 2} & \multicolumn{4}{c}{Average} \\ \cmidrule(lr){2-5} \cmidrule(lr){6-9} \cmidrule(lr){10-13}
 & CNR & MSR & TP & EP & CNR & MSR & TP & EP & CNR & MSR & TP & EP \\ \midrule
Patch Ensemble (CNR) & \textcolor{red}{12.09} & \textcolor{blue}{3.55} & \textcolor{brown}{1.10} & \textcolor{red}{0.74} & \textcolor{red}{13.12} & \textcolor{brown}{4.25} & \textcolor{brown}{1.07} & \textcolor{blue}{0.75} & \textcolor{red}{12.60} & \textcolor{blue}{3.90} & 1.09 & \textcolor{blue}{0.74} \\ 
Patch Ensemble (TP) & 11.62 & 3.35 & \textcolor{red}{1.19} & \textcolor{blue}{0.73} & 12.68 & 3.97 & \textcolor{blue}{1.10} & \textcolor{blue}{0.75} & 12.15 & 3.66 & \textcolor{blue}{1.14} & \textcolor{blue}{0.74} \\ 
Patch Ensemble (MSR) & \textcolor{brown}{11.93} & \textcolor{red}{3.59} & \textcolor{brown}{1.10} & \textcolor{red}{0.74} & \textcolor{blue}{13.04} & \textcolor{red}{4.40} & 1.06 & \textcolor{blue}{0.75} & \textcolor{blue}{12.49} & \textcolor{red}{3.99} & \textcolor{brown}{1.08} & \textcolor{blue}{0.74} \\ 
Patch Ensemble (EP) & 11.62 & 3.44 & 1.01 & \textcolor{red}{0.74} & 12.44 & 4.00 & 0.95 & \textcolor{red}{0.76} & 12.03 & \textcolor{brown}{3.72} & 0.98 & \textcolor{red}{0.75} \\ 
\textbf{Patch Ensemble (All)} & \textcolor{blue}{11.96} & \textcolor{brown}{3.53} & \textcolor{blue}{1.18} & \textcolor{red}{0.74} & \textcolor{brown}{12.92} & \textcolor{blue}{4.28} & \textcolor{red}{1.12} & \textcolor{blue}{0.75} & \textcolor{brown}{12.44} & \textcolor{blue}{3.90} & \textcolor{red}{1.15} & \textcolor{blue}{0.74} \\ 
\bottomrule

\end{tabular}
}
\end{table}

\begin{figure}[h]
    \centering
    \includegraphics[width=\linewidth]{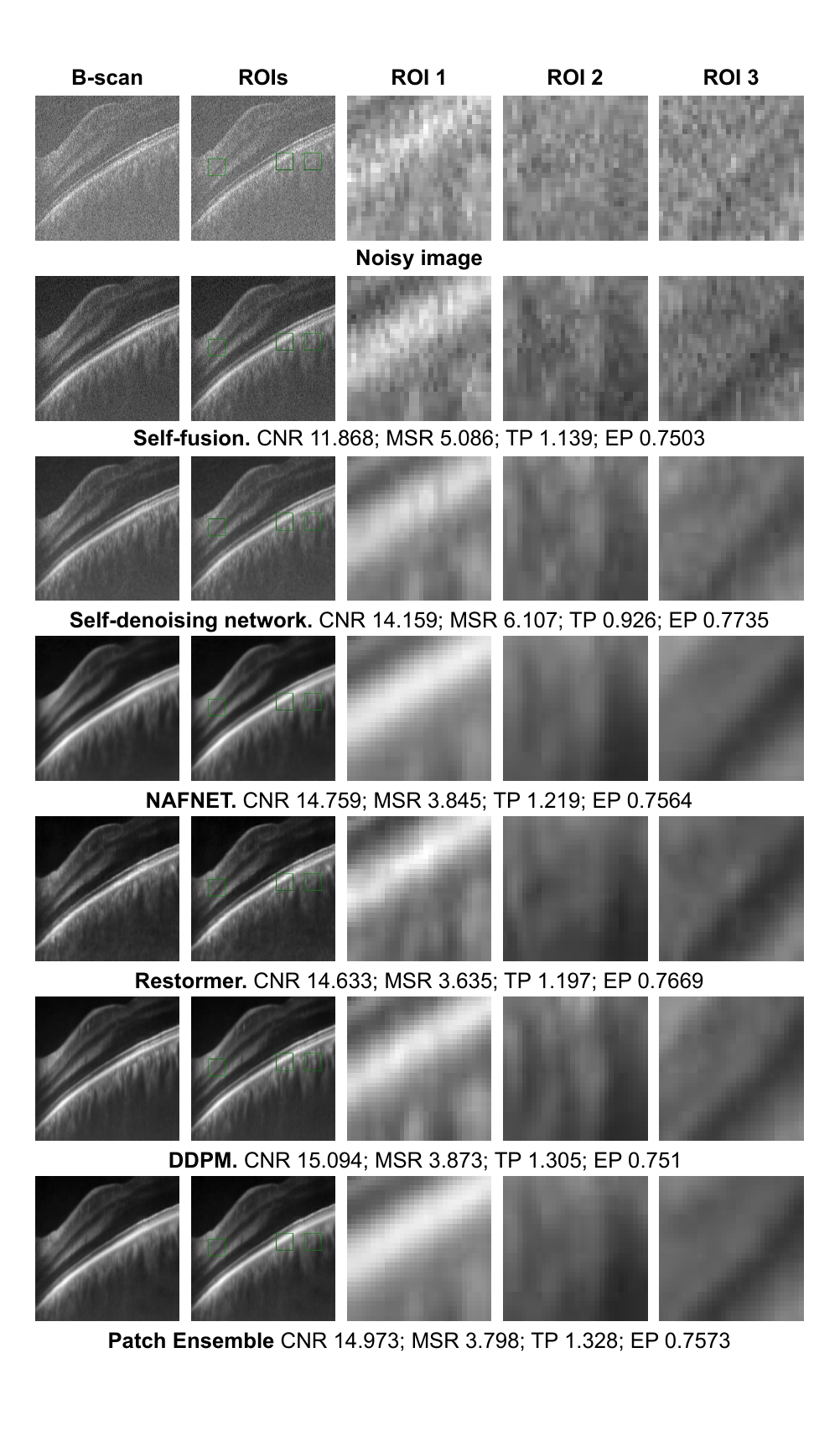}
    \vspace{-4em}
    \caption{Visual results of used methods on a B-scan}
    \label{fig:visual}
\end{figure}

Figure \ref{fig:visual} shows the denoised images of a challenging input noise sample.
% TODO: write discussions for visual results
\vspace{-1.5em}

\section{Conclusion}
We introduced a novel two-phase denoising pipeline for OCT images, leveraging self-fusion, neighbor-based self-supervision, and a patch-based ensemble of
advanced denoising networks. Our framework tackles the common challenge in
medical image denoising: the unavailability of perfectly clean ground truth data.
By smartly creating pseudo-clean references and aggregating the strengths of
multiple models, our method robustly suppresses speckle noise and preserves crit-
ical anatomical details. Experimental results on the noisy-only VIP Cup dataset
demonstrate the effectiveness of our approach, paving the way for improved OCT
image quality and more reliable clinical diagnoses.

\color{red}\section*{Ablation Study}
\begin{table}[h!]
\centering
\vspace{-1em}
\caption{Self-fusion adaptation difference}
\setlength{\tabcolsep}{4pt} % Adjust column separation
\scalebox{0.82}{
\begin{tabular}{lcccccccc}
\toprule
 & \multicolumn{4}{c}{\textbf{Fold 1}} & \multicolumn{4}{c}{\textbf{Fold 2}} \\ 
\cmidrule(lr){2-5}\cmidrule(lr){6-9}
\textbf{Method} & \textbf{CNR} & \textbf{MSR} & \textbf{TP} & \textbf{EP} & \textbf{CNR} & \textbf{MSR} & \textbf{TP} & \textbf{EP} \\
\midrule
Self-fusion                & 10.80 & 4.70 & 0.93 & 0.83 & 10.49 & 4.91 & 0.94 & 0.84 \\
R2R\cite{GR2R}      & 8.231 & 4.55 & 0.98 & 0.78 & 7.65 & 4.60 & 0.97 & 0.78 \\
Self-fusion + R2R       & 10.13 & 4.55 & 1.08 & 0.67 & 9.89 & 4.78  & 1.08 & 0.67 \\
Neighbor2Neighbor   & 10.63 & 10.36 & 0.47 & 0.87 & 9.95 & 10.38 & 0.47 & 0.87 \\
Self-fusion + Neighbor2Neighbor    & 13.46 & 5.90  & 0.73 & 0.84 & 13.20 & 6.28  & 0.74 & 0.84 \\
\bottomrule
\end{tabular}
}
\end{table}

\begin{table}[h!]
\centering
\vspace{-1em}
\caption{COLOR SCALING ADAPTATION DIFFERENCE}
\setlength{\tabcolsep}{4pt} % Adjust column separation
\scalebox{0.9}{
\begin{tabular}{lcccccccc}
\toprule
 & \multicolumn{4}{c}{\textbf{Fold 1}} & \multicolumn{4}{c}{\textbf{Fold 2}} \\ 
\cmidrule(lr){2-5}\cmidrule(lr){6-9}
\textbf{Method} & \textbf{CNR} & \textbf{MSR} & \textbf{TP} & \textbf{EP} & \textbf{CNR} & \textbf{MSR} & \textbf{TP} & \textbf{EP} \\
\midrule
Scaled CGNet   & 13.966 & 4.360 & 1.017 & 0.835 & 13.841 & 4.569 & 1.019 & 0.837 \\
CGNet  & 13.946 & 5.220 & 0.726 & 0.844 & 13.822 & 5.748 & 0.700 & 0.845 \\
Scaled NAFNET   & 14.542 & 3.776 & 1.159 & 0.822 & 14.573 & 4.277 & 1.103 & 0.832 \\
NAFNET  & 14.531 & 6.177 & 0.728 & 0.844 & 14.561 & 6.973 & 0.721 & 0.847 \\
Scaled Restormer   & 14.427 & 3.963 & 1.026 & 0.822 & 14.108 & 4.118 & 0.977 & 0.829 \\
Restormer  & 14.452 & 7.494 & 0.539 & 0.847 & 14.134 & 7.710 & 0.523 & 0.849 \\
\bottomrule
\end{tabular}
}
\label{table:colortable}
\end{table}

% TODO: write discussions for visual results
\vspace{-3.5em}
\begin{figure}[h]
    \centering
    \includegraphics[width=\linewidth]{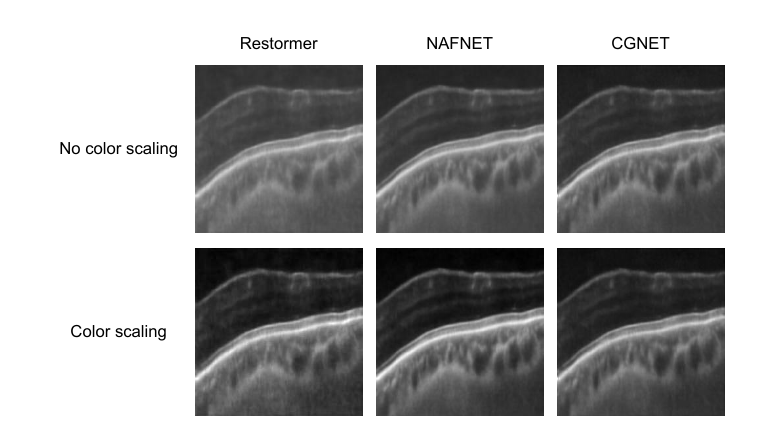}
    \vspace{-2.7em}
    \caption{Visual of color scaling method}
    \label{fig:colorab}
\end{figure}

\begin{table}[h!]
\centering
\vspace{-1em}
\caption{Performance Comparison of Different Methods. The best top 3 results are \textcolor{red}{Red}, \textcolor{blue}{Blue}, \textcolor{brown}{Brown}}
\setlength{\tabcolsep}{6pt}
\scalebox{0.9}{
\begin{tabular}{lcccc}
\toprule
\textbf{Method} & \textbf{CNR} & \textbf{MSR} & \textbf{TP} & \textbf{EP} \\
\midrule
Scaled CGNET            & 13.966 & 4.360 & 1.017  & \textcolor{red}{0.835} \\
Scaled NAFNET             & \textcolor{blue}{14.542} & \textcolor{red}{3.776} & \textcolor{blue}{1.159}  & 0.822 \\
Scaled Restormer             & 14.427 & 3.963 & 1.026  & 0.822 \\
Patch Ensemble (CNR)                 & 14.381 & 3.977 & 1.061  & 0.824 \\
Patch Ensemble (MSR)                  & 14.102 & \textcolor{blue}{4.320} & 1.021  & 0.833 \\
Patch Ensemble (TP)                  & 13.463 & \textcolor{brown}{3.728} & \textcolor{brown}{1.180}  & 0.822 \\
Patch Ensemble (EP)                   & 13.967 & 4.205 & 1.043  & \textcolor{blue}{0.834} \\
Patch Ensemble (cnr1-msr3-tp10)     & 13.930 & 4.183 & 1.130  & \textcolor{brown}{0.828} \\
Patch Ensemble (cnr3-msr2-tp5)     & 14.050 & 4.075 & 1.080  & 0.826 \\
Patch Ensemble (cnr1-msr1-tp10)      & 13.777 & 4.021 & 1.149  & 0.825 \\
Patch Ensemble (cnr3-tp5)           & 14.291 & 3.943 & 1.097  & 0.824 \\
Patch Ensemble (cnr1-tp5)            & 14.111 & 3.867 & 1.1450 & 0.822 \\
Patch Ensemble (cnr1-tp20)         & 13.868 & 3.758 & 1.178  & 0.822 \\
Voting Ensemble             & \textcolor{brown}{14.479} & 3.859 & 1.101  & 0.824 \\
Model inference             & \textcolor{red}{14.601} & 3.832 & \textcolor{red}{1.221}  & 0.819 \\
\bottomrule
\end{tabular}
}
\label{table:ablation_methods}
\end{table}
Notation: 

\textcolor{black}{Patch Ensemble (cnr1-msr3-tp10) =  CNR*1 + MSR*3 + TP*10.}

\end{document}